\documentclass{cta-author}

\usepackage{algorithmic}
\usepackage[ruled,norelsize]{algorithm2e}

{}
{}
{}

\usepackage{enumitem}
\setitemize[0]{leftmargin=10pt,itemindent=10pt,listparindent=5pt,topsep=1pt}

\begin{document}

%\supertitle{Submission for IET Research Journal Papers}

\title{An Efficient UAV-based Artificial Intelligence Framework for Real-Time Visual Tasks}

\author{\au{Enkhtogtokh Togootogtokh$^{1\corr}$}, \au{Christian Micheloni$^{2}$}, \au{Gian Luca Foresti$^{2}$}, \au{Niki Martinel$^{2}$}
%\au{Sunan Huang$^{3}$}, \au{Wai Lun Leong$^{3}$}, \au{Swee Huat Rodney Teo$^{3}$}
}

\address{\add{1}{AViReS Lab, University of Udine and Mongolian University of Science and Technology, Viale Delle Scienze, 206, Udine, Italy}
\add{2}{AViReS Lab, University of Udine, Viale Delle Scienze, 206, Udine, Italy}
%\add{3}{Temasek Lab, National University of Singapore, Singapore}
\email{enkhtogtokh.java@gmail.com}}

\begin{abstract}
Modern Unmanned Aerial Vehicles equipped with state of the art artificial intelligence (AI) technologies are opening to a wide plethora of novel and interesting applications.
While this field received a strong impact from the recent AI breakthroughs, most of the provided solutions either entirely rely on commercial software or provide a weak integration interface which denies the development of additional techniques.
This leads us to propose a novel and efficient framework for the UAV-AI joint technology.
Intelligent UAV systems encounter complex challenges to be tackled without human control.
One of these complex challenges is to be able to carry out computer vision tasks in real-time use cases.
In this paper we focus on this challenge and introduce a multi-layer AI (MLAI) framework to allow easy integration of ad-hoc visual-based AI applications.
To show its features and its advantages, we implemented and evaluated different modern visual-based deep learning models for object detection, target tracking and target handover. 
\end{abstract}

\maketitle

\section{Introduction}\label{sec1}
Unmanned Aerial Vehicles (UAVs) have great potential to be widely used in real-life applications.
Because of their low cost, safety benefit and mobility, UAVs can potentially replace manned aerial vehicles in many tasks as well as perform very well in tasks that tradition manned aerial vehicles do not. 
This is especially true when UAVs are equipped with modern state-of-the-art AI technologies delivering astonishing performance in many challenging tasks (\emph{e.g.}, computer vision~\cite{CV}, natural language processing~\cite{NLP}, etc.)
Within this setting, despite the recent efforts, modern AI implementations are still missing due to the difficulties of real-time exchanging and control of the data shared between the UAV and the intelligent part of the system.
This generally deny an easy development of ad-hoc AI solutions. 

We propose an efficient and flexible multi layer AI (MLAI) framework which has been conceived considering the entire AI pipeline from sensor data reading to results delivery.
We borrowed the long term software development technologies such as the quite common front-end and back-end methodology for our AI framework. For the front-end layer, a desktop user interface is implemented with the help of the object oriented C\# and XAML technologies. For the middle communication layer, we setup a smooth socket message exchange to let a rapid communication between other layers. For the back-end layer, the python 3.x platform is adopted and frameworks such as pytorch, numpy, and opencv-python are exploted for AI.
To control the UAV flight, we implemented an Android application using the commercial DJI SDK. 
To demonstrate the benefits of the proposed MLAI framework, we have selected the visual re-identification task~\cite{Martinel2018a}. 
This requires completion of the visual object detection, tracking and handover sub-tasks to be successfully completed.
We have considered the recent breakthrough of deep neural networks (DNN) to tackle such tasks.

The major feature DNNs is the ability to learn visual features automatically. This contrasts with the feature-engineering approach where previous domain knowledge is required to come up with a suitable feature representation. 
Local receptive fields, shared weights, pooling, and sub sampling are the main concepts within traditional convolutional neural networks (CNNs). These help to handle target translations, scale variations, or distortions. 
In industry, the system needs to employ state-of-the-art technologies to meet customer  requirements \cite{IET}.
For this research, we implemented two top state-of-the-art AI algorithms as YOLOv3 \cite{YOLOv3} and DCFNet \cite{DCFNet} to detect and track objects from UAV, respectively.  
YOLOv3 is one of the best algorithms for real-time object detection. Three versions have already been developed with significant continuous improvements. YOLOv3 is the latest version, which we implemented in this research. This algorithm is $1.5\times$ or $2\times$ faster than other solutions such as ResNet-101\cite{DeepRes} or ResNet-152\cite{DeepRes}.
For visual object tracking, current state-of-the-art algorithms are founded on the Discriminative Correlation Filter (DCF) technique.
Such a technique gives optimal real-time performance when multi-channel features can be exploited.
It is a matter of fact that better features always increase tracking performance.
Today the trend within this research topic is to use the multi-layer deep features for DCF tracking to improve the tracking performances. Among such methods, DCFNet \cite{DCFNet} is a very light weight method achieving state-of-the-art performance.
Due to this reason, it has been selected and implemented for real-time visual object tracking.
Finally, to conclude the re-identification, we implemented a new two-way handover approach which grounds on the previous work done in~\cite{Martinel2018a}.

\noindent Concretely, the key contributions of the proposed work are:
\begin{itemize}
\item A novel and efficient architecture for visual-based AI system.
\item Real-time state-of-the-art deep learning implementations.
\item A two-way handover technique.
\item Definition of minimal internet of drone things.
\end{itemize}

\noindent Systematic experiments conducted on real-world data have shown that the proposed solution: 
\begin{itemize}
\item is able to detect multiple objects with a processing speed of 19.65 frames per second (fps).
\item can track an object of interest in real-time (at 29.94 fps).
\item is able to improve the one-way handover performance by a significant margin (about $30\%$ accuracy improvement).
\end{itemize}

\noindent The rest of the paper is organized as follows.
The proposed MLAI framework is described in Section~\ref{proposedarch}.
The internet of drone things is explained in Section~\ref{iodt}.
The details about the implemented deep learning algorithms and the handover process are discussed in Section \ref{deeplearning} and \ref{reid}, respectively. 
The experimental results are presented in Section \ref{experimentalresult}.
Finally, Section \ref{conclusion} provides the conclusions and future work.

\section{The MLAI Framework}\label{proposedarch}
\begin{figure}[t]
  \centering
  \includegraphics[width=\linewidth]{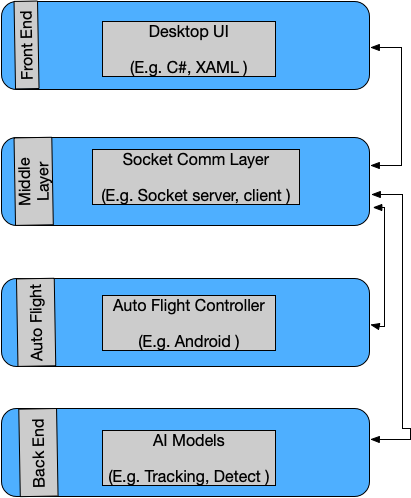}
  \caption{Pipeline of the proposed MLAI framework. The first layer is for UI, the second layer is for communication, third layer is for automatic flight control, and final layer is for AI models.}
  \label{fig:system_arch}
\end{figure}

In the followin, we present the details of the proposed MLAI framework for drone-related AI applications.
The overall framework pipelines is shown in Figure~\ref{fig:system_arch}.
The framework architecture has four layers which have been inspired from the long term web application development, hence considering a front- and back-end solution. 
The first layer is for front-end development, while the middle layer is in charge of managing the communication between the front-end and back-end layers.
The third layer consists of the drone flight controlling activities, while the fourth one is the back-end layer which is in charge of carrying out the AI activities.
This is achieved by designing deep learning models through pytorch, numpy, opencv-python and additional utility libraries.
The framework has the following main advantages:
\begin{itemize}
    \item  Portable. The front-end and the back-end layers might reside in different locations and places within the network environment. 
    \item  Feasible. Since AI frameworks are mostly developed through the usage of python platforms such that PyTorch~\cite{PyTorch}, TensorFlow~\cite{TensorFlow}, and python itself, we don't need develop another framework to use in different programming language. 
    \item Multiple user interfaces. If additional user interfaces are needed (e.g., a user interface running on a mobile application), just adding a new simple socket connection is enough.
    \item Agile development and testing. Front-end and back-end developers can make suitable changes without worrying about dependencies, thus working independently and in parallel.  
\end{itemize}

It is worth to note that, while it has been primarily designed for drone systems, the proposed MLAI framework can been seen as a general architecture for any AI implementations where real-time and easy exchange of data between a sensor and the intelligent part of the system is needed.

\subsection{Front-end UI layer}
\begin{figure}[h]
  \centering
  \includegraphics[width=\linewidth]{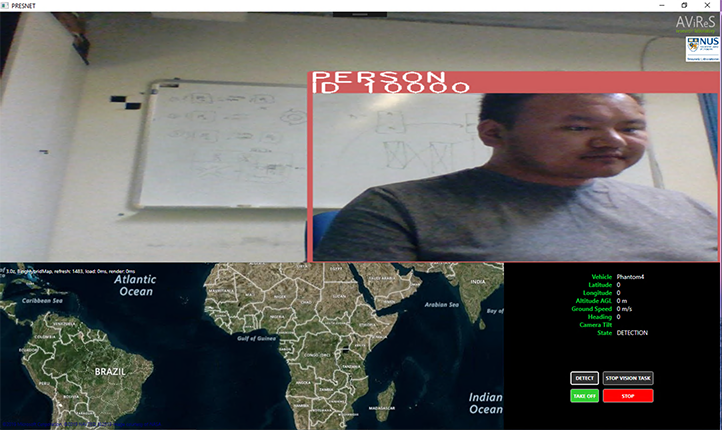}
  \caption{Front-end UI Application. Realtime Video Stream and Inference Results are on top panel, The visual navigation map is on the bottom left panel, and controlling panel is on the bottom right panel.}
  \label{fig:front-end}
\end{figure}

The front-end layer consists of main user interface (UI) which has three sub components: (i) the real-time video panel, (ii) the map panel, (iii) and the user control panel.
It is a desktop application designed following the model-view-controller pattern and implemented by using the XAML, C\#, GMap.NET, and OpenCVSharp framework technologies.
The main user interface is shown in Figure \ref{fig:front-end}.
The top video streaming panel shows the video footages acquired from the flying sensor, and adds any inference result that is delivered by the back-end AI layer through the middle communication socket layer. The left bottom map panel shows the UAV current position and has been implemented by using the GMap.Net component. The bottom right control panel shows the current status of the UAV by listing the vehicle model, latitude, longtude, altitude, ground speed, heading, camera tilt, and state information.
Start and stop computer vision task control button as well as UAV automatic take off and stop control buttons are also present.    

\subsection{Middle communication layer}
Here we discuss in detail more about socket handlers and how we can efficiently design the middle connection layer.
We develop in this layer socket servers and clients which are handling video signal from front-end to back-end layer, exchange the messages as inference results, text messages, and commands. 
The important solution is to transfer the real-time video signal without latency and packet loss. Since UDP socket is the promising to have low latency however no nature for packet loss, we use TCP socket directly to smoothly handle the video signal without latency and packet loss. We develop algorithm to encode and decode the matrix of video frame. The algorithm \ref{alg:matencode} is to perform the encoding matrix of video signal. With utility of OpenCV ImEncode function it takes the frame as input then encodes to bytes to transfer through socket to back-end AI layer.
\begin{algorithm}[!t]
\caption{Matrix Encode} \label{alg:matencode}
\begin{algorithmic}[1]

\STATE using OpenCvSharp ;
\STATE using System.Net.Sockets ;
\texttt{\\}
\texttt{\\}
\STATE byte[]        bytes ;
\STATE Mat           frame ;
\STATE SocketClient  socketClient ;
\texttt{\\}
\texttt{\\}
\STATE Initialize() ;
\IF {!frame.Empty())}
\STATE    Cv2.ImEncode(".jpg",frame, out bytes); 
\STATE    socketClient.SendByte(bytes); 
\ENDIF

\end{algorithmic}
\end{algorithm}

It is useful to directly put the syntax here as three members of class are initialized and then opencv ImEncode function is directly applied to encode the frame matrix to send it to back-end AI layer.   
\begin{algorithm}[h]
\caption{Matrix Decode} \label{alg:matdecode}
\begin{algorithmic}[1]
\STATE import cv2
\STATE import numpy as np
\texttt{\\}
\texttt{\\}
\STATE data = socket.recv(BUFFER\_SIZE)
\STATE array = np.fromstring(data,np.uint8)
\STATE frame = cv2.imdecode(array, cv2.IMREAD\_COLOR)
 
\end{algorithmic}
\end{algorithm}

In Algorithm \ref{alg:matdecode},  we use numpy library and opencv python since our AI models are working in python platform. To decode the frames, first, encoded matrix is converted into array then we applied opencv imdecode function to decode the matrix.

\subsection{Back-end AI layer}
\begin{figure}[h]
  \centering
  \includegraphics[width=1\linewidth]{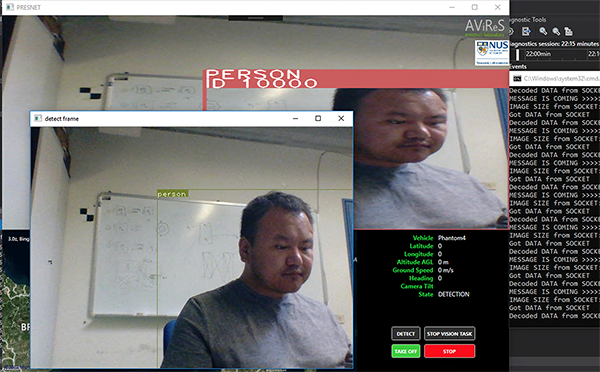}
  \caption{Back-end AI platform. Debugging windows and logging terminal with front-end UI. }
  \label{fig:back-end}
\end{figure}

In this section, we develop python platform with utilities from numpy, opencv-python, and PyTorch frameworks as shown in Figure \ref{fig:back-end}. Main AI approaches are state-of-the-art algorithms as YOLO and DCFNet approaches.  All state-of-the-art deep leaning algorithms discussed in Section \ref{deeplearning}.  It receives the video signal through socket from front-end layer and then once decodes the signal it does input that signal into AI models in well order. For real-time system, since our AI algorithms work in low resolution frame, it is always better to reduce the size of video frame. It receives all signals from front-end layer as video signal and object detect, track, and vision task stop command strings through communication middle socket layer. Then it sends back  the information of object detection and tracking bounding box, recognized object labels, accuracy, and executing command result strings. 
To debug AI tasks we use two visual windows for object recognition and tracking algorithms as shown in Figure \ref{fig:back-end}.  Once front-end desktop application starts, we then run the back-end python platform up as well using shell calling API in C\# technology. 

\subsection{Auto flight control for real-time AI tasks}
We implement the auto flight control for real-time AI tasks in Android platform based on DJI Mobile SDK. DJI products have their auto flight control API with mission and virtual stick controls. To implement real-time AI tasks, the virtul stick control (VSC) is the reliable approach. And, we propose our own auto pilot control application based on VSC API.  Camera gimbal module was running in DJI SDK standard function. However, it was not functioning as expected in real-time integration tests. We improved it as individual safe thread to function in real-time integration case. 
In more detail, we have four important components such that throttle, yaw, pitch, and roll controlling (TYPRC) component, gimbal tilt controlling component, telemetry data handling component, and communication socket TCP component.   The TYPRC is based on DJI automatic flight controller API.
 In generally, the flight controller is responsible as following:
\begin{itemize}
    \item Flight control including motor control, taking off, and landing
    \item Aircraft state information such that attitude, position, and speed
    \item Sensor sub components such that compasses, positioning systems
    \item Aircraft sub components such that the landing gear
    \item Aircraft flight simulation for testing and debugging
\end{itemize} 

Specially, we implemented the DJI Virtual Stick Control which is automated the aircraft fly in flexible and the best solution for computer vision intelligent tasks compare to another automation mission control. It commands to move the aircraft horizontally with X/Y velocities or roll, pitch angles. Larger roll and pitch angles make in larger X and Y velocities, respectively. Roll and pitch directions are dependent on the coordinate system. 
The telemetry component provides state information handling at up to 10 Hz including:
\begin{itemize}
    \item Aircraft position, velocity and altitude
    \item Home location
    \item Sensor information
    \item Whether motors are on or off
    \item Flight limitation and GEO system information
\end{itemize} 

\begin{figure}[h]
  \includegraphics[width=\linewidth]{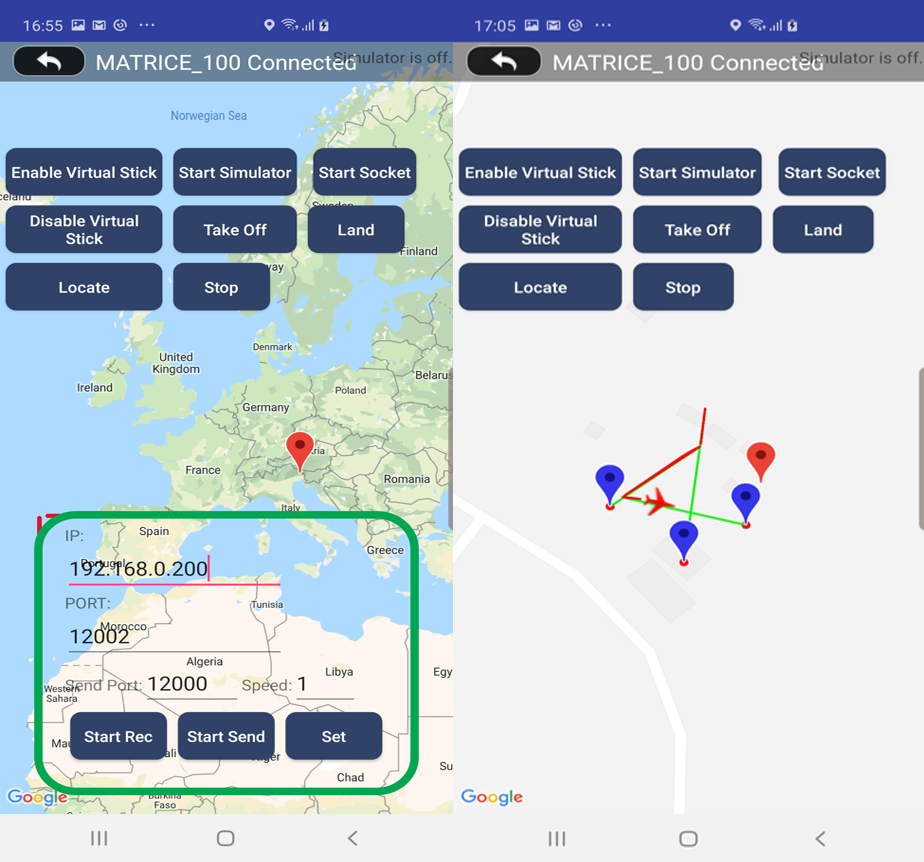}
  \caption{Automatic Flight Control Component Testing UI. The left window is to configuration for socket connection. And right window is to visualize the drone flight and control buttons UI.}
  \label{fig:autoflight}
\end{figure}

In user interface (UI), the application has the map as google API, communication controller (UI), and automatic flight controlling UI.   
To visualize the aircraft motion, google map is implemented in our application as show in Figure \ref{fig:autoflight}. We tested the aircraft yaw, pitch, roll, and throttle controlling functionality as using this component as well.

The most important UAV auto heading algorithm is provided in Algorithm \ref{alg:autoheading}. In Algorithm \ref{alg:autoheading}, we assume that UAV target is to approach from A point to B point. To perform this task, first we have compute bearing angle then convert it into the quad copter angle system. 

\begin{algorithm}
\caption{UAV Auto Heading} \label{alg:autoheading}
\begin{algorithmic}[1]

\STATE LatLng a; 
\STATE LatLng b;
\texttt{\\}
\texttt{\\}
 
\STATE //Calculate bearing
\STATE double lng1,lng2,lat1,lat2,
\STATE        lngDiff,x,y,bearing;
\STATE  lng1 = a.longitude;
\STATE  lng2 = b.longitude;
\STATE  lat1 = Math.toRadians(a.latitude);
\STATE  lat2 = Math.toRadians(b.latitude);
\STATE  lngDiff= Math.toRadians(lng2-lng1);
\STATE  y= Math.sin(lngDiff)*Math.cos(lat2);
\STATE  x=Math.cos(lat1)
\STATE    *Math.sin(lat2)-Math.sin(lat1)
\STATE    *Math.cos(lat2)*Math.cos(lngDiff);
\STATE  bearing = (Math.toDegrees(
\STATE           Math.atan2(y, x))  +360)%360;
\STATE //Calculate quad copter angle system 
\STATE //[-180;180]
\IF {bearing == 0}
\STATE     return 0;
\ENDIF
\STATE int d = bearing/180;
\IF  {d\%2 == 0}
      \RETURN angle\%180;
\ENDIF 
\STATE int signum = Math.abs(angle)/angle;
\STATE int translatedAngle = angle\%180 - signum*180;
\STATE return translatedAngle;
 
\end{algorithmic}
\end{algorithm}

In the next section, we will discuss what minimal internet of things shall work to bring modern AI for our UAV.

\subsection{The Internet of Drone Things (IoDT) }\label{iodt}
\begin{figure}[h]
  \centering
  \includegraphics[width=0.8\linewidth]{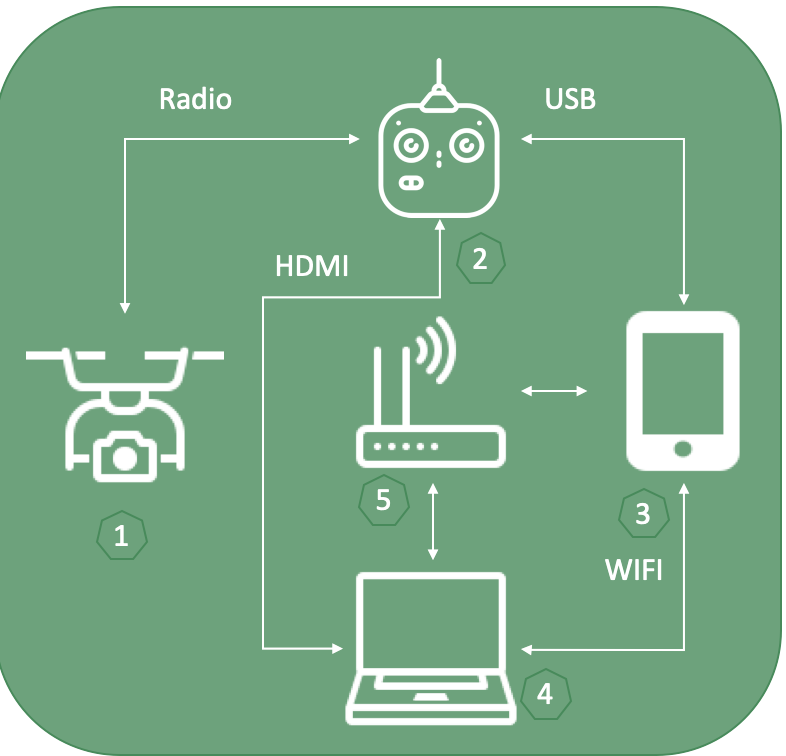}
  \caption{Minimal Internet of Drone Things setup. All connection types are visualized from (1) to (4).}
  \label{fig:iodt}
\end{figure}

In this section, we define what internet of drone things need to be connected to achieve the main goal as to bring AI ability for the drone (see Figure \ref{fig:iodt}):
\begin{itemize}
\item The Drone. 
\item Remote Control.
\item Smart Phone.
\item Laptop.
\item WIFI Router.
\end{itemize}
As shown in Figure \ref{fig:iodt}, the drone and remote controller (RC) connect through wireless, RC and mobile device communicate through USB cable, and mobile device and laptop connect to the same WIFI router. To stream live video from the drone, we directly use HDMI port between laptop and RC. It was the simple and most effective solution.      
In generally for standard case like today's most distributed and programmable drone industry availability, it is the optimal solution as the minimum number of connected devices to have the AI ability. 
Here we only discuss about in terms of connection of the things. The rest of whole setup and more details are given in Section \ref{experimentalresult}.
To exchange data through connection, we mostly use low level network programming as TCP sockets. 
 
\subsection{Deep Neural Networks for Drone}\label{deeplearning}
We implemented two strong deep neural network models for our UAV to track and recognize the objects. Both tracking and detection approaches are given in detail based original researches in \cite{YOLOv3}, \cite{DCFNet}. 

\subsubsection{Target Detection and Recognition}
An artificial neural network architecture is used to perform the feature extraction. Specifically, the adopted network follows a hybrid model inheriting from YOLOv2 \cite{Yolo9000}, Darknet-19 and the modern residual networks. 
The network has 53 convolutional layers with some shortcut connections. It is mainly composed of 3x3 and 1x1 convolutional layers.  
The goal of the detection task is to predict the width and height as well as as the location of a bounding box enclosing an object of interest. In YOLOv3 \cite{YOLOv3}, these are the coordinates $t_x, t_y, t_w, t_h$.

In the following  are the width and height of bounding box priors. Being $\sigma$ a sigmoid function and considering the cell offset from the top-left corner of the image by $(c_x,c_y)$, then a bounding box prediction corresponds to:
\begin{equation}\label{eq_det_1}
    b_x=\sigma(t_x)+c_x, 
    b_y=\sigma(t_y)+c_y,
    b_w=p_we^t_w, 
    b_h=p_he^t_h
\end{equation}

Let $\hat{t}_*$ be the ground truth for some coordinate prediction and let $t_*$ be the predicted value, then the gradient can be computed as $\hat{t}_*-t_*$.  We can compute the  by inverting the above equations. The model uses logistic regression to predict the objectness score for each bounding box.  
It predicts the classes of the bounding box which may contain multiple labels. Since the softmax has been shown to be not much relevant for good performance, it has not been included within the model. Hence, independent logistic regression classifiers have been used. For class predictions, the binary cross-entropy is implemented.

\subsubsection{Target Tracking}
For DCFNet tracking appraoch, the conv1 from VGG \cite{VGG} is only the convolutional layers for the lightweight network with 75KB. The output is forced to 32 channels and all pooling layers are removed. Stochastic Gradient Descent (SGD) algorithm is applied to train the network with momentum 0.9, weight decay  and the learning rate.
Twenty epoch is looped for the model with mini-batch size of 16.  The online learning rate is fixed as 0.008 for the hyper-parameter in the correlation filter layer. We fixed the Gaussian spatial bandwith as 0.1 and the regularization coefficient as 1e-4. The patch pyramid is used with the scale factors.

\subsection{Re-Identification (ReID) Handover Process }
\label{reid}
\subsubsection{Two-Way Re-Identification}
\begin{figure}[!t]
  \centering
  \includegraphics[width=\linewidth]{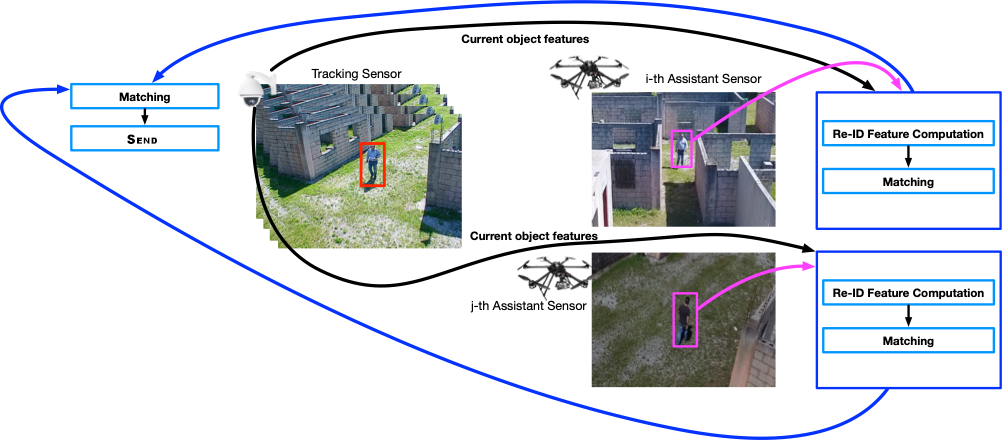}
  \caption{Two-way Re-Identification pipeline. The current tracking object is surrounded by a red bounding box. Objects detected by each assistant sensor are depicted in magenta}
  \label{fig:reid1}
\end{figure}

The proposed ReID scheme is shown in Figure \ref{fig:reid1}. The whole process starts with a camera/UAV tracking the target of interest. We will refer to such a camera/UAV as the tracking sensor. During the tracking procedure the tracking sensor acquires images of the target (under different illumination conditions and poses) and computes the ReID features which are stored, together with the target ID, within the tracking sensor gallery. The assistant cameras/UAVs (dubbed assistant sensors) which are asked to support the current tracking sensor will receive the current tracking target feature representation. After this is received, they run the proposed detection and centroid tracking solutions one after the other. This is necessary to assign the same ID (within each assistant sensor list) to the objects that are appearing in their field-of-view (FoV) at different time instants. For each of such objects, the feature representation is computed and added to the gallery list of objects belonging the assistant sensor. After such an inclusion, the received target feature representation $x$ is compared with the feature representation $y$ of each image in the gallery list of the assistant by using the cosine similarity

\begin{equation}
 cos(x,y) = \frac{x^Ty}{||x|||y||}
\label{eq:eq1}
\end{equation}
which returns a matching score representing the ReID confidence. If the match is above a threshold (currently set to $60\%$ chance of matching) the features obtained from the new objects are sent back to the tracking sensor.
This runs the second ReID match by using the feature representation received from the assistant sensor as the new probe. This is matched against the gallery list of objects belonging to the tracking sensor, which includes the current tracking object. The same cosine similarity is used as the matching metric. After the match with all gallery objects is concluded, a ranking list of candidates is obtained by sorting the matched gallery objects in descending order. We then count how many times the tracking target falls in the top-k matches of such a ranked list (we currently set $k=20$). We use such a counter ($z$) together with the number of times a specific object is received from the assistant drone ($t$) to weight the matching score as 
\begin{equation}
\phi(x,y,z,t)  = ztcos(x,y)=zt \frac{x^Ty}{||x|||y||}
\label{eq:eq2}
\end{equation}

Such a score is finally sent back to the core system which is in charge of deciding to which assistant sensor the tracking sensor has to handover. This process is repeated until the tracking sensor receives the stop ReID signal and the selected assistant sensor receives the start tracking command. This completes the handover.

\subsubsection{Target Feature Representation}
\begin{figure}[!t]
  \centering
  \includegraphics[width=\linewidth]{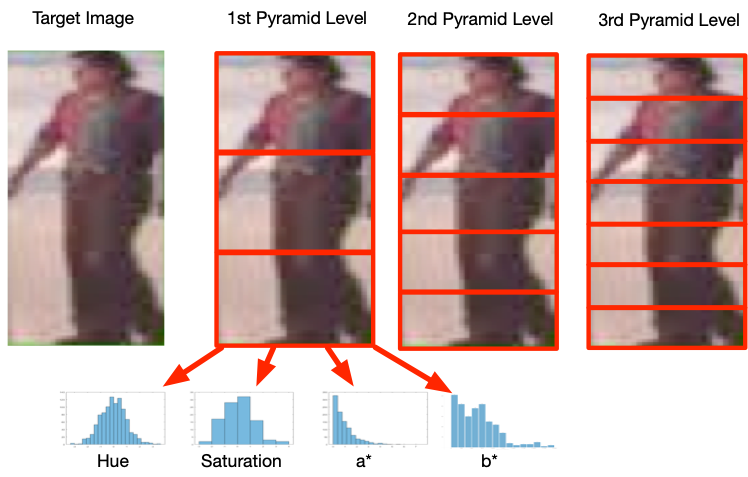}
  \caption{ReID feature description computation. Histogram feature extraction is carried out for each of the stripes dividing the target image at the different levels of the pyramid structure}
  \label{fig:reid2}
\end{figure}

To have a more robust solution to illumination changes and pose variations, in this last milestone, we have also introduced a new target representation (shown in Figure \ref{fig:reid2}). 

This leverages on the image pyramid framework to compute multi-level features for image stripes, which allow us to be more robust to left/right viewpoint changes and misaligned tracking results. We considered three pyramid levels with a split of the person into 3 different stripes at the first level. At the 2nd and 3rd levels of the pyramid, the target image is divided into 5 and 7 stripes, respectively. For each of the image stripes 4 different histograms are extracted from the Hue, Saturation, and a* and b* components of the Hue-Saturation-Value and CIELa*b* color spaces. Histograms computed for all stripes are then concatenated to provide the final feature representation.
Then obtained vector is finally normalized by using the power norm law followed by the L2 normalization.

\section{Experimental Results}\label{experimentalresult}
In this section, we discuss first about the setup, and then evaluate the efficiency of the MLAI framework, finally detection, tracking, and handover object ReID results are experimented in systematic scenarios. 
\subsection{Setup}
\begin{figure}[h]
  \centering
  \includegraphics[width=\linewidth]{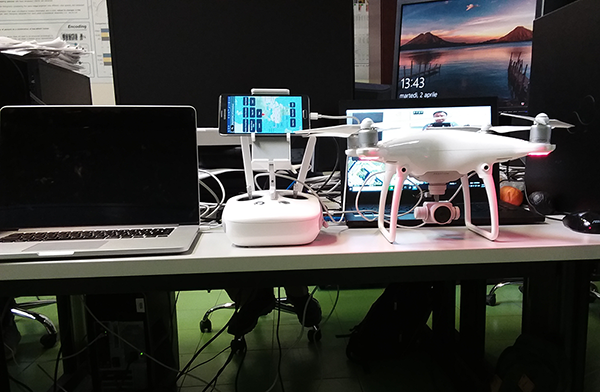}
  \caption{The setup with Phantom 4, it's RC, Android device, and GPU powered laptop.}
  \label{fig:setup}
\end{figure}

We setup the DJI Phantom 4, Matrice-100, their standard remote controllers, 2 Samsung Galaxy Note 4 mobile devices (CPU: Quad Core SnapDragon, 2.7GHz, RAM: 3GB), and 2 Asus Laptops (Republic of Gamers, CPU:Core i7-875H@2.20GHz, 2.21GHz, RAM:16GB, GPU: NVidia GeForce GTX 1070, 8 GB) as shown in Figure \ref{fig:setup}.  
\subsection{The MLAI Framework Efficiency}
In this section, we evaluate the simple efficiency experimental results for the MLAI framework. The software performance index Apdex score is the industry standard for tracking relative performance of the software. Since our final product that using the MLAI is the software. Apdex works by specifying a goal for how long a specific request should take. In our case requests are mostly to detect and track that specific objects. Those requests are then classified into satisfied (fast enough),tolerating, not satisfied (too slow), and failed. The math formula is as:
\begin{equation}\label{eq_apdex}
    ApdexScore=\frac{C_s+\frac{C_t}{2}}{C_{total}} 
\end{equation}
where $C_s, C_t,C_{total}$ are satisfied count, tolerating count, total samples, respectively. 
In Table \ref{tab:efficiency}, we experimented 100 samples to request as track and detect the specific objects with threshold t=0.5 sec and then according Apdex scores are computed.
\begin{table}[h]
  \caption{The MLAI Framework Efficiency}
  \label{tab:efficiency}
  \begin{tabular}{ccccc}
   
      \hline
      Request & $C_s$ & $C_{t}$ & $C_{total}$& Apdex Score  \\
      \hline
    
     Object Detect & 97 & 3& 100 & $0.985 $  \\
     Object Track & 98 & 2 & 100& $0.99$  \\

  \end{tabular}
\end{table}

\subsection{Detection and Recognition}
To evaluate the performance of the selected YOLOv3 detector we have acquired footages from UAV camera. In the following sections we report on the performance achieved considering the Scenario 1 - person detection, Scenario 2 - vehicle detection, and Scenario 3 - multi objects detection, respectively.
\subsubsection{Scenario 1 - Person Detection}
\begin{figure}[h]
  \includegraphics[width=\linewidth]{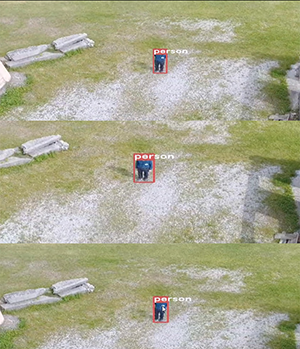}
  \caption{Scenario 1 - Person Detection. The person is walking on demonstration site.}
  \label{fig:detect_sc1}
\end{figure}

We experimented the results on the person detection case as shown in Figure \ref{fig:detect_sc1}. A person was walking on the testing area and we collected here some frame sequences from the video. As we can see the algorithm works in real-time and continuously detecting.

\subsubsection{Scenario 2 - Vehicle Detection}
\begin{figure}[h]
  \includegraphics[width=\linewidth]{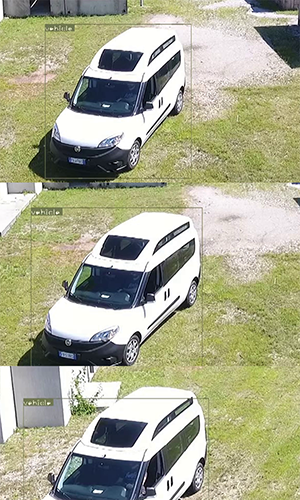}
  \caption{Scenario 2 - Vehicle Detection. The vehicle is running on demonstration site.}
  \label{fig:detect_sc2}
\end{figure}

We experimented the results on the vehicle detection case as shown in Figure \ref{fig:detect_sc2}. The drone is following the vehicle on the test area and we collected here some frame sequences from the video. As we can see the algorithm works real-time and continuously detecting the vehicle.

\subsubsection{Scenario 3 - Multiple Objects Detection}
\begin{figure}[h]
  \includegraphics[width=\linewidth]{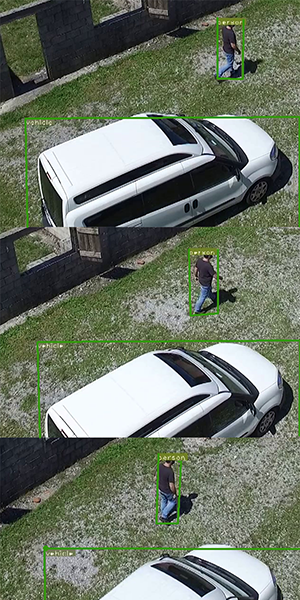}
  \caption{Scenario 3 - Multiple Objects Detection. The person and vehicle are moving on testing site.}
  \label{fig:detect_sc3}
\end{figure}

 In Figure \ref{fig:detect_sc3}, the drone is following the vehicle and person on the testing area and we collected here some frame sequences from the video. As we can see the algorithm works real-time and continuously detecting the vehicle and person both.

\subsection{Tracking}
In this section, we experiment the tracking results. To evaluate the tracking solution, we have collected multiple videos at the test-bed site. Specifically, we have acquired footages simulating the evaluations scenarios.

\subsubsection{Scenario 1 - Person Tracking}
\begin{figure}[h]
  \includegraphics[width=\linewidth]{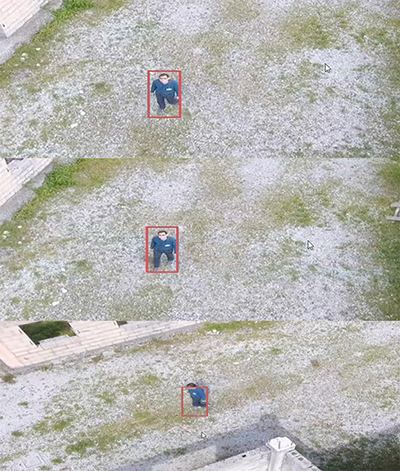}
  \caption{Scenario 1 - Person Tracking}
  \label{fig:tracking_sc1}
\end{figure}

We report on the achieved results on the person tracking case as shown in Figure \ref{fig:tracking_sc1}. A person was walking on the testing area. Results show that real-time performance are achieved with stable tracking results.
\subsubsection{Scenario 2 - Vehicle Tracking}
\begin{figure}[t]
  \includegraphics[width=\linewidth]{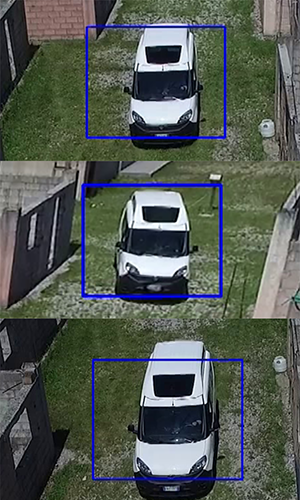}
  \caption{Scenario 2 - Vehicle Tracking. The vehicle is driving on test site with normal speed.}
  \label{fig:tracking_sc2}
\end{figure}

 In Figure \ref{fig:tracking_sc2}, the drone is following the vehicle on the testing area and we collected here some frame sequences from the video. As we can see the algorithm works real-time and continuously tracking the vehicle.

\subsubsection{Scenario 3 - Multi Target Tracking}
\begin{figure}[t]
  \includegraphics[width=\linewidth]{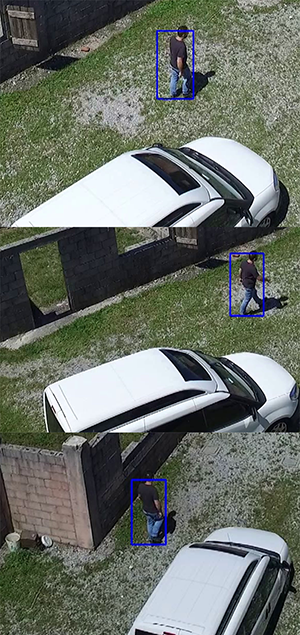}
  \caption{Scenario 3 - Muti Targets Tracking. The person is selected to track in multiple different objects.}
  \label{fig:tracking_sc3}
\end{figure}

We experimented the results on the multiple objects as vehicle and person tracking case as shown in Figure \ref{fig:tracking_sc3}. The drone is following the vehicle and person on the testing area and we collected here some frame sequences from the video.

\subsection{Re-identification}
\begin{figure}[t]
  \includegraphics[width=\linewidth]{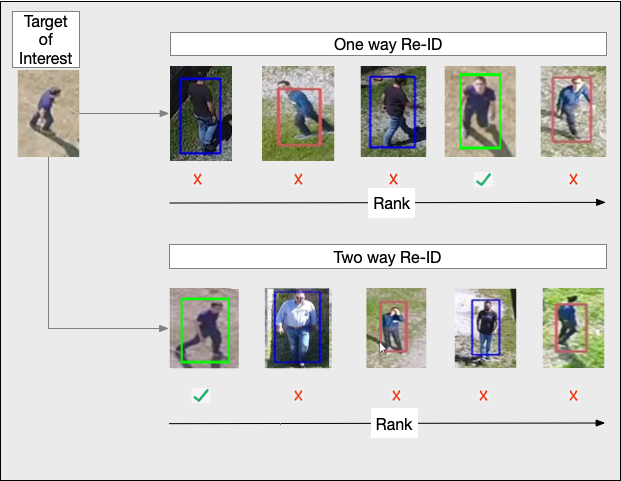}
  \caption{Two-way ReID experimental results. The first result is one way ReID. Second result is two way ReID. Correct match has a tick on its bottom, wrong ones have a cross. }
  \label{fig:reid3}
\end{figure}

Here we experiment the re-identification results. To evaluate the re-identification solution, we have collected multiple videos at the test-bed site. Specifically, we have acquired footages simulating the evaluations scenarios. Please note here our proposed two way ReID approach has about 30\% higher accuracy than one way ReID approaches. We show the experimenting results in Figure \ref{fig:reid3}. 

\section{Conclusion}\label{conclusion}
We proposed the efficient and novel MLAI general framework and implemented it for UAV AI implementation. Modern state-of-the-art hybrid deep learning approaches implemented to recognize and track to bring AI for UAV in real-time to provide the novel and complete approach for object re-identication. And then the re-identification handover process is introduced in this research especially for UAV. We defined minimal IoDT for UAV. All main important algorithms are directly provided in this research to develop the novel UAV system powered by AI.

\vfill\pagebreak

\end{document}